\documentclass[sigplan,screen]{acmart}

\AtBeginDocument{%
  \providecommand\BibTeX{{%
    \normalfont B\kern-0.5em{\scshape i\kern-0.25em b}\kern-0.8em\TeX}}}

\setcopyright{acmcopyright}
\copyrightyear{2018}
\acmYear{2018}
\acmDOI{10.1145/1122445.1122456}

\acmConference[Computing Frontiers '21]{Computing Frontiers '21: ACM International Conference on Computing Frontiers}{May 11--13, 2021}{Virtual Conference}
\acmBooktitle{Computing Frontiers '21: ACM International Conference on Computing Frontiers,
  May 11--13, 2021, Virtual Conference}
\acmPrice{15.00}
\acmISBN{978-1-4503-XXXX-X/18/06}

\usepackage[redeflists]{IEEEtrantools}
\newcommand{\sm}{\text{{SpiNeMap}}}{}
\newcommand{\pc}{\text{{PyCARL}}}{}
\newcommand{\esl}{\text{{Decompose\-SNN}}}{}
\usepackage{amsmath}
\usepackage{multirow}
\usepackage{gensymb}
\usepackage{xcolor}
\usepackage[caption=false]{subfig}
\usepackage[linesnumbered,ruled]{algorithm2e}

\SetCommentSty{mycommfont}
\SetKwRepeat{Do}{do}{while}
\usepackage{threeparttable}

\usepackage{pifont}

\newcommand{\ineq}[1]{\footnotesize$#1$\normalsize}{}
\newtheorem{Tlemma}{Lemma}

\newtheorem{Tdef}{Definition}
\newenvironment{Definition}[1]
    {\begin{Tdef}\noindent\textsc{(#1)}\itshape}
    {\end{Tdef}}

\begin{document}
\bstctlcite{IEEEexample:BSTcontrol}
\title{On the Role of System Software in Energy Management of Neuromorphic Computing}

\author{Twisha Titirsha}
\email{tt624@drexel.edu}
\authornote{Authors contributed equally to this research.}
\affiliation{%
  \institution{Drexel University}
  \city{Philadelphia}
  \state{PA}
  \country{USA}
  \postcode{19104}
}
\author{Shihao Song}
\authornotemark[1]
\email{ss3695@drexel.edu}
\affiliation{%
  \institution{Drexel University}
  \city{Philadelphia}
  \state{PA}
  \country{USA}
  \postcode{19104}
}

\author{Adarsha Balaji}
\authornotemark[1]
\email{ab3586@drexel.edu}
\affiliation{%
  \institution{Drexel University}
  \city{Philadelphia}
  \state{PA}
  \country{USA}
  \postcode{19104}
}

\author{Anup Das}
\email{anup.das@drexel.edu}
\affiliation{%
  \institution{Drexel University}
  \city{Philadelphia}
  \state{PA}
  \country{USA}
  \postcode{19104}
}

\begin{abstract}
  Neuromorphic computing systems such as DYNAPs and Loihi have recently been introduced to the computing community to improve performance and energy efficiency of machine learning programs, especially those that are implemented using Spiking Neural Network (SNN). 
  The role of a system software for neuromorphic systems is to cluster a large machine learning model (e.g., with many neurons and synapses) and map these clusters to the computing resources of the hardware. In this work, we formulate the energy consumption of a neuromorphic hardware, considering the 
  power consumed by neurons and synapses, and the energy consumed in communicating spikes on the interconnect. Based on such formulation, we first evaluate the role of a system software in managing the energy consumption of neuromorphic systems. Next, we formulate a simple heuristic-based mapping approach to place the neurons and synapses onto the computing resources to reduce energy consumption. We evaluate our approach with 10 machine learning applications and demonstrate that the proposed mapping approach leads to a significant reduction of energy consumption of neuromorphic computing systems.
\end{abstract}




\maketitle

\section{Introduction}\label{sec:introduction}
Neuromorphic computing describes the VLSI implementation of the neuro-biological architecture of the central nervous system~\cite{mead1990neuromorphic,catthoor2018very,balaji2019design}. 
Neuromorphic systems are energy efficient in executing Spiking Neural Networks (SNNs), which are considered as the third generation of neural networks~\cite{maass1997networks}. SNNs use spike-based computations and bio-inspired learning algorithms in solving machine learning problems.
In an SNN, pre-synaptic neurons communicate information encoded in spike trains to post-synaptic neurons, via the synapses.
Performance of an SNN-based application can be assessed in terms of the inter-spike interval (ISI coding) or mean firing rate of the neurons (rate coding).

\begin{figure}[h!]
	\centering
	\vspace{-5pt}
	\centerline{\includegraphics[width=0.99\columnwidth]{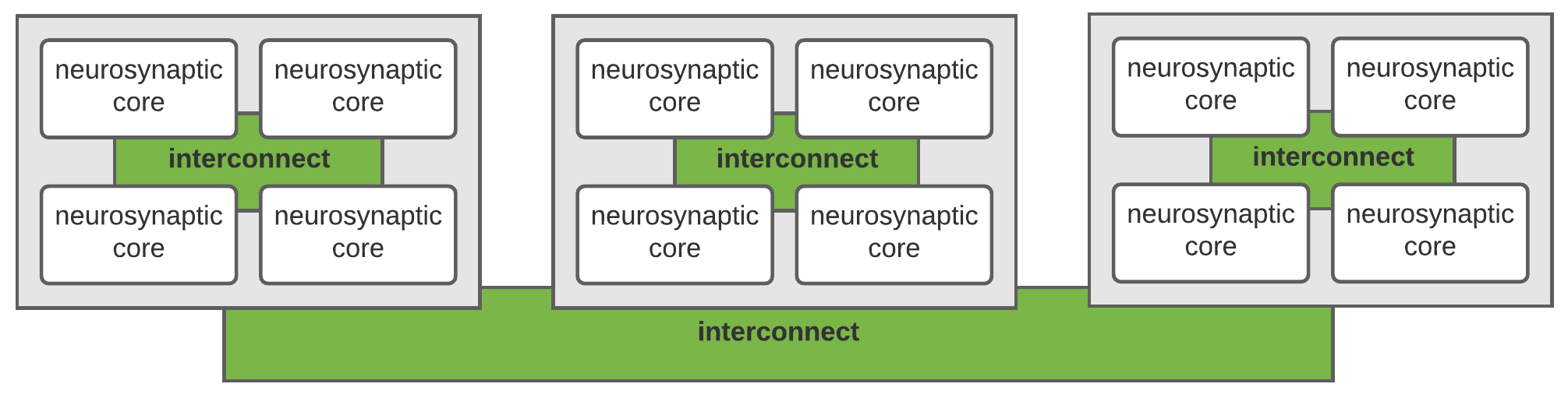}}
	\vspace{-10pt}
	\caption{Hardware architecture of neuromorphic systems.}
	\vspace{-10pt}
	\label{fig:hardware}
\end{figure}

The hardware architecture of neuromorphic systems consists of neurosynaptic cores,
which are interconnected via a shared interconnect~\cite{balaji2019exploration}. Figure~\ref{fig:hardware} illustrates the representative hardware architecture of many recent neuromorphic systems such as Loihi~\cite{loihi}, TrueNorth~\cite{truenorth}, and DYNAPs~\cite{dynapse}.

Neurosynaptic cores are the computing resources of a neuromorphic system. In many emerging architectures, a neurosynaptic core is essentially a crossbar array that can accommodate a fixed number of neurons and synapses. The role of a system software for neuromorphic systems is therefore, to partition an SNN with many neurons and synapses into clusters such that,
the neurons and synapses of each cluster can be mapped on to a neurosynaptic core of the system.
Therefore, the inter-cluster synapses are mapped on the shared interconnect of the system.

Recently, energy consumption of neuromorphic systems has come into spot light, specifically due to their application in power-constrained environments such as Embedded Systems and Internet-of-Things (IoT). 
To this end, several low-power analog neuron designs are proposed to implement neurosynaptic cores for neuromorphic systems~\cite{indiveri2003low,woo2020implementation,rubino2020ultra,yang2020analog,natarajan2018hodgkin}. Another research direction is the shift from Static RAM (SRAM)-based synapse designs to implementations that use Non-Volatile Memory (NVM) -- Phase Change Memory (PCM)~\cite{Burr2017}, Oxide-based Resistive RAM (OxRRAM)~\cite{mallik2017design}, Ferroelectric RAM (FeRAM)~\cite{mulaosmanovic2017novel}, and Spin-Transfer Torque Magnetic or Spin-Orbit-Torque RAM (STT/ SoT-MRAM)~\cite{vincent2015spin}.\footnote{Beside neuromorphic computing, some of these memristor technologies are also used as main memory in conventional computers to improve performance and energy efficiency~\cite{palp,mneme,datacon,hebe,shihao_igsc}.}

In addition to the hardware-oriented energy reduction techniques, we argue that the system software also plays a pivotal role in the energy consumption of neuromorphic systems. We show that the energy consumption of a neuromorphic hardware depends on 1) how an SNN model is partitioned into clusters, 2) how the clusters are mapped to the neurosynaptic cores, and 3) how the neurons and synapses of a cluster are placed inside each core. Following are our key \textbf{contributions}.
\begin{itemize}
    \item We formulate the energy consumption of a neuromorphic hardware, considering the energy consumed inside each neurosynaptic core and the energy consumed in communicating spikes on the shared interconnect.
    \item We show that by not considering all the sources of energy loss, existing system software approaches leave a significant energy improvement opportunities.
    \item We propose a heuristic to minimize the total energy consumption in neuromorphic computing without significantly increasing the spike latency. This leads to only a very marginal impact on performance.
    \item We evaluate our mapping approach with 10 machine learning workloads on a cycle-accurate simulator of a state-of-the-art neuromorphic hardware.
\end{itemize}

Results demonstrate that the current system software frameworks for neuromorphic systems miss significant energy improvement opportunities.
By explicitly incorporating energy consumption of different hardware units, the proposed mapping approach significantly minimizes the energy consumption of neuromorphic systems.

\section{Background}\label{sec:background}
There are many recent initiatives to map machine learning workloads to neuromorphic hardware. PACMAN~\cite{galluppi2015framework} is used to map SNNs to SpiNNaker hardware~\cite{spinnaker}. Corelet~\cite{amir2013cognitive} is used to map workloads to TrueNorth~\cite{truenorth}. NEUTRAM~\cite{ji2016neutrams} is a mapping approach for digital neuromorphic chips such as TIANJI~\cite{tianji}. 
PyNN~\cite{pynn}, which started as a front-end to many back-end SNN simulators such as Brian~\cite{brian}, NEURON~\cite{neuron}, and NEST~\cite{nest}, can now map SNN applications to many neuromorphic hardware such as Loihi~\cite{loihi} and Neurogrid~\cite{neurogrid}. 
A recent extension of PyNN, called PyCARL~\cite{pycarl}, can simulate SNN applications using the back-end CARLsim simulator~\cite{carlsim}, allowing mapping of these applications to the DYNAPs neuromorphic hardware~\cite{dynapse}.
There are also other propritary approaches to mapping SNN applications to emerging neuromorphic chips such as BrainScaleS~\cite{brainscale}, Braindrop~\cite{braindrop}, and ODIN~\cite{odin}. 

DecomposedSNN~\cite{esl20} uses spatial decomposition technique to unroll each neuron with many fanin connections into smaller atomic units that are connected sequentially. This allows to densely pack each crossbar in a neuromorphic hardware leading to a significant improvement of resource utilization and a reduction of hardware area overhead.
PSOPART~\cite{psopart} and SpiNeMap~\cite{spinemap} are mapping approaches that minimize spike communication energy on the shared interconnect by lowering the spike volume and spike latency, respectively.
SPINERTM~\cite{balaji2020run} is proposed to remap SNN applications to neuromorphic hardware at tun-time by monitoring the performance degradation.
DFSynthesizer~\cite{dfsynthesizer} uses data flow models to analyze performance of SNN workloads on crossbar-based neuromorphic hardware. There are also other dataflow-based technique reported in literature~\cite{balaji2019ISVLSIframework,das2018dataflow,adarsha_igsc}. 
These approaches are demonstrated with many SNN applications, such as the liquid state machine (LSM)-based heart-rate estimation~\cite{HeartEstmNN}; spiking ResNet architecture for ImageNet classification~\cite{sengupta2019going}; deep learning architecture for DNA sequence analysis~\cite{moyer2020machine}; heart-rate classification using spiking CNN architecture using ECG data~\cite{jolpe18,das2018heartbeat}; lateral inhibition-based digit recognition~\cite{Diehl2015}; recurrent architecture-based predictive visual pursuit~\cite{Kashyap2018}; spiking architecture for seizure classification using EEG data~\cite{ghosh2007improved}; among others.

RENEU~\cite{reneu} is a recent technique proposed to map SNN applications to hardware improving the circuit aging of the peripheral circuitry in crossbars, which is caused due to their high-voltage exposure. There are also other approaches targeting circuit aging~\cite{song2020case,balaji2019framework}.
ESPINE~\cite{espine} is an approach to map SNN applications to neuromorphic hardware, improving the endurance of its Non-Volatile Memory synapses. There are also other mapping approaches that target temperature optimization~\cite{twisha_thermal} and releiability-performance trade-offs~\cite{twisha_endurance,vts_das}.
We compare our Hill Climbing approach against \pc{} and \sm{}, and found it to perform significantly better in terms of energy consumption.


\section{Problem Formulation}\label{sec:operational_semantics}

\begin{figure}[h!]
	\centering
	\vspace{-5pt}
	\centerline{\includegraphics[width=0.99\columnwidth]{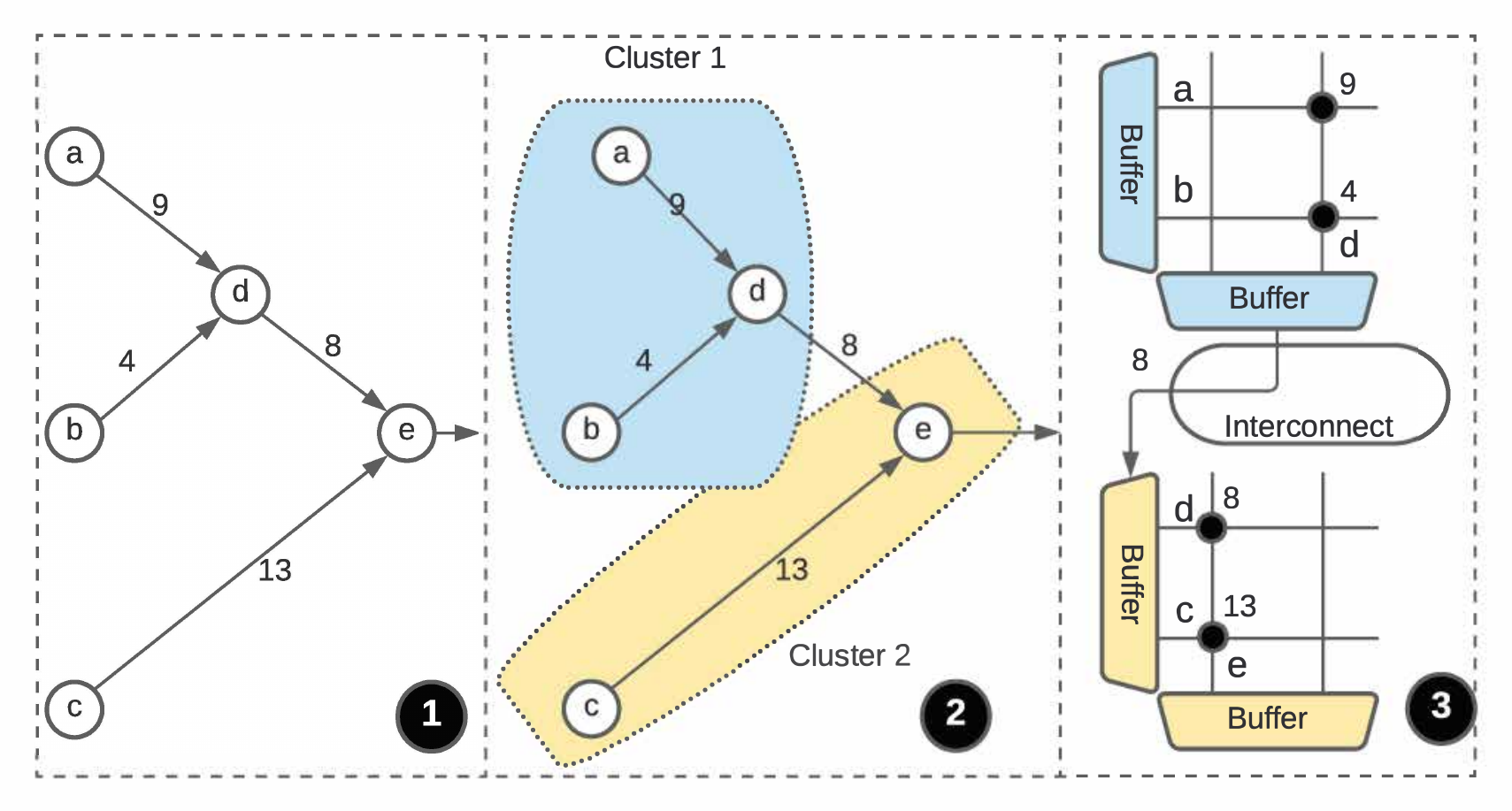}}
	\vspace{-10pt}
	\caption{A typical clustering approach used by system software to map SNNs to neuromorphic hardware.}
	\vspace{-10pt}
	\label{fig:clustering_demo}
\end{figure}

Unlike system software for conventional computers (e.g., the Operating System), the role of the system software for neuromorphic hardware is to cluster a machine learning model and map the clusters onto the crossbars of the neuromorphic hardware.
Figure~\ref{fig:clustering_demo} illustrates the mapping concept using an example SNN shown in (\ding{182}). The number on a link represents the average number of spikes communicated between the source and destination neurons for a representative training data. We consider the mapping of this SNN to 
a hardware with $2 \times 2$ crossbars. Since a crossbar in this hardware can only accommodate a maximum of 2 pre-synaptic connections, we partition the SNN of (\ding{182}) into two clusters (shown in two different colors) in (\ding{183}). These clusters can then be mapped to the two crossbars as shown in (\ding{184}), with an average 8 spikes communicated between the crossbars.

In many neuromorphic applications, the number of pre-synaptic connections per neuron can well exceed the crossbar input limit (which is typically 128 or 256). For those applications, each neuron is first decomposed into smaller units with fewer pre-synaptic connections before they are clustered using the approach illustrated in Fig.~\ref{fig:clustering_demo} (see~\cite{esl20}).

\begin{figure}[h!]
	\centering
	\vspace{-5pt}
	\centerline{\includegraphics[width=0.59\columnwidth]{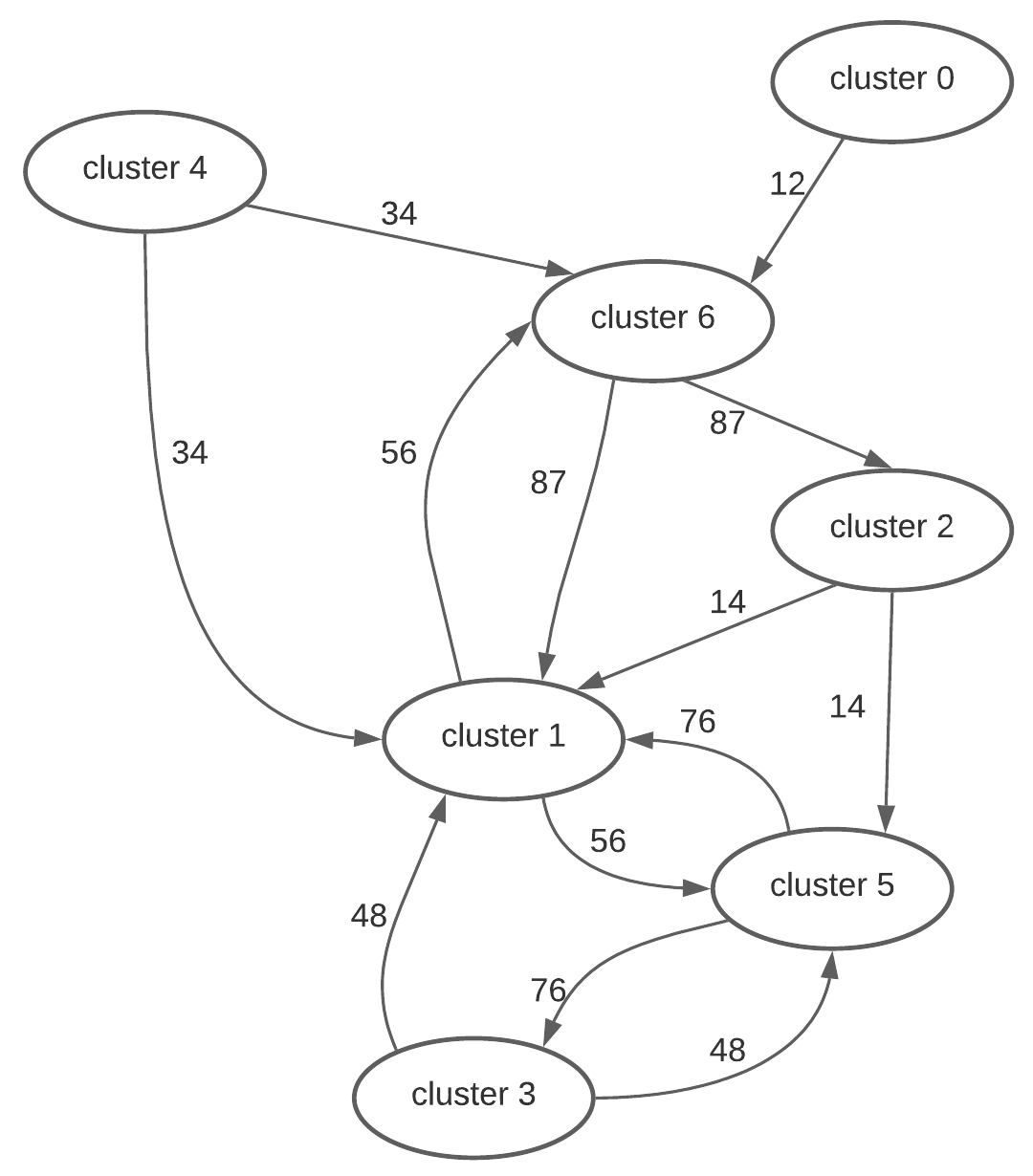}}
	\vspace{-10pt}
	\caption{The clusters generated from LeNet CNN.}
	\vspace{-10pt}
	\label{fig:cluster_lenet}
\end{figure}

Figure~\ref{fig:cluster_lenet} illustrates the clusters 7 clusters of the LeNet Convolutional Neural Network (CNN) obtained using the clustering technique of \sm{}.

Formally, a clustered SNN graph is defined as follows.
\begin{Definition}{Clustered SNN}
A clustered SNNs \ineq{\mathbf{G_{CSNN} = (\textbf{C},\textbf{L})}} is a directed graph consisting of a finite set \ineq{{\textbf{C}}} of clusters and a finite set \ineq{{\textbf{L}}} of connections between these clusters.
\end{Definition}

Each cluster \ineq{C_i\in\textbf{C}} is a tuple \ineq{\langle In(C_i), Out(C_i),S(C_i),W(C_i)\rangle}, where \ineq{In(C_i)} is the number of pre-synaptic neurons of the cluster, \ineq{Out(C_i)} is the number of post-synaptic neurons of the cluster, \ineq{S(C_i)} is the number of spikes generated inside the cluster, and \ineq{W(C_i)} is the set of synaptic weights of the cluster. Each link \ineq{L_i\in \textbf{L}} of the graph has a value \ineq{Spk(L_i)} attached to it representing the number of spikes communicated on the link between the source and the destination clusters.

The clusters of an SNN-based application are mapped to the tiles of a neuromorphic hardware, where a tile consists of a neurosynaptic core, e.g., a crossbar. 

Formally, a neuromorphic hardware is defined as follows.

\begin{Definition}{{Neuromorphic Hardware}}
	 A neuromorphic hardware \ineq{G_{NH} = (\textbf{T},\textbf{I})} is a directed graph consisting of a finite set \ineq{\textbf{T}} of tiles and a finite set \ineq{\textbf{I}} of interconnect links. 
\end{Definition}

Each tile consists of a crossbar to map neurons and synapses, and input and output buffers to receive and send spikes over the interconnect, respectively. A tile \ineq{T_i\in \textbf{T}} is a tuple \ineq{\langle M,InB(T_i),OutB(T_i)\rangle}, where \ineq{M} is the dimension of a crossbar on the tile, i.e., the tile \ineq{T_i} can accommodate \ineq{M} pre-synaptic neurons, \ineq{M} post-synaptic neurons, and \ineq{M^2} synaptic connections, \ineq{InB(T_i)} is the input buffer size on the tile, and \ineq{OutB(T_i)} is its output buffer size. Each interconnect link is bidirectional, representing two-way communication between the source and destination tiles with a fixed bandwidth \ineq{BW}.

When mapping the clusters to the tiles of the hardware, spikes from a tile (i.e., the cluster mapped to the tile) are broadcasted on the interconnect. The network interface (NI) logic on each tile ensures the delivery of the spikes to the intended recipient neurons mapped to these tiles.

To formalize the energy consumption, we consider the mapping of a clustered SNN \ineq{\mathbf{G_{CSNN} = (\textbf{C},\textbf{L})}} to the neuromorphic hardware \ineq{G_{NH} = (\textbf{T},\textbf{I})}. 

Mapping \ineq{M:\mathbf{G_{CSNN} = (\textbf{C},\textbf{L})}\rightarrow G_{NH} = (\textbf{T},\textbf{I})} is specified by a logical matrix \ineq{(m_{ij}) \in \{0,1\}^{|\textbf{C}|\times|\textbf{T}|}}, where \ineq{m_{ij}} is defined as
\begin{equation}
    \label{eq:mapping_rep}
    \footnotesize m_{ij} = \begin{cases}
    1 & \text{if cluster } {C}_i \in \textbf{C} \text{ is mapped to tile } {T}_j\in\textbf{T}\\
    0 & \text{otherwise}
    \end{cases}
\end{equation}

To simplify the discussion, we consider a neuromorphic hardware to have as many tiles as clusters of a given application. The energy formulation also holds when the tiles are time-multiplexed among the clusters~\cite{dfsynthesizer}.

\section{Energy Modeling}\label{sec:energy_model}
In this section, we provide a comprehensive energy model for neuromorphic hardware executing machine learning workloads.
We consider the following energy components.
\subsection{Spike Energy}
This is the total energy consumed on the tiles to generate all the spikes for a given SNN application working on a representative data.

Using the formalism of Section~\ref{sec:operational_semantics}, the spike energy is
\begin{equation}
    \footnotesize E_{spk} = \sum_{i=0}^{|\textbf{C}|-1} \cdot \sum_{j=0}^{S(C_i)-1} e_{spk}(i,j),
\end{equation}
where \ineq{e_{spk}(i,j)} is the energy of \ineq{j^\text{th}} spike on tile \ineq{C_i}. Since each cluster is mapped to a tile of the hardware, the outer summation is for all the clusters of an application, while the inner summation is for all the spikes generated inside each cluster. 
\ineq{e_{spk}(i,j)} comprises of two components -- the energy to generate a spike by a pre-synaptic neuron (\ineq{e_{neuron}}), which remains the same for all the tiles (ignoring process variation for the moment), and the energy consumed on a synaptic cell due to the flow of current (\ineq{e_{synapse} (i,j)}). Therefore,
\begin{equation}
    \footnotesize e_{spk}(i,j) = e_{neuron} + e_{synapse}(i,j).
\end{equation}
In all previous work, the energy per spike is typically assumed to be constant. However, here we show this is not the case.
In general, the synaptic energy depends on the specific NVM used to model the synaptic weights in a neurosynaptic core. We formulate this for the Phase-Change Memory (PCM). 
%
The \textbf{scope} of this work is on the inference phase, wherein a machine learning model is pre-trained offline, and the trained model is programmed on the neuromorphic hardware. Therefore, we focus on the energy to read the synaptic weights stored in the PCM cells of a crossbar.

The energy consumed in propagating current through a PCM cell is given by Joule Heating~\cite{espine,das2014reinforcement,das2013reliability,das2015reliability,das2014temperature,das2016adaptive}
\begin{equation}
    \label{eq:e_synapse_1}
    \footnotesize e_{synapse} = I_{prog}^2\cdot \left(R_{synapse} + R_{ON}\right)\cdot t_{spk},
\end{equation}
where \ineq{I_{prog}} is the current generated for the spike voltage (\ineq{\approx 50\mu A}), \ineq{R_{synapse}} is the resistance of the PCM cell, \ineq{R_{ON}} is the ON resistance of the access transistor connecting the PCM cell and \ineq{t_{spk}} is the spike duration (typically a few ms). Considering \ineq{W(C_i)} to be the synaptic weights of the cluster \ineq{C_i}, which are programmed as conductances, Eq.~\ref{eq:e_synapse_1} can be written as
\begin{equation}
    \label{eq:e_synapse_1}
    \footnotesize e_{synapse}(i,j) = I_{prog}^2\cdot t_{spk}\cdot\left(R_{ON} + \frac{1}{w(i,j)}\right),
\end{equation}
where \ineq{w(i,j)\in W(C_i)} is the conductance of the PCM cell on the path of the \ineq{j^\text{th}} spike in the cluster \ineq{C_i}.

\begin{figure}[h!]
	\begin{center}
		\vspace{-10pt}
		\includegraphics[width=0.39\columnwidth]{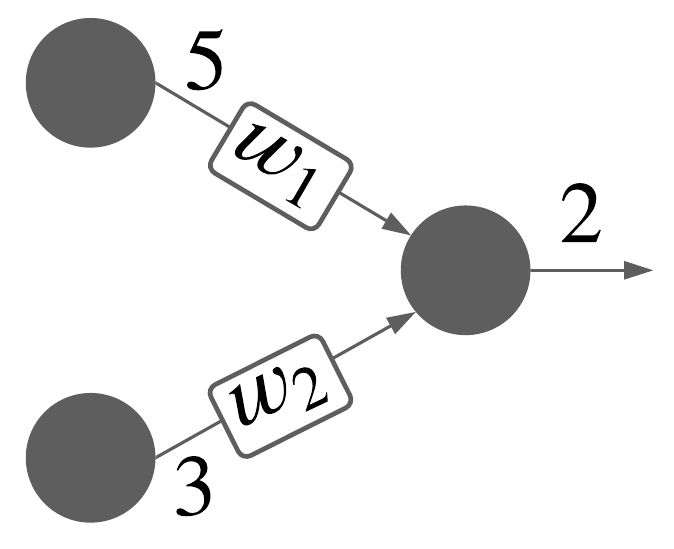}
		\vspace{-10pt}
		\caption{Example of calculating \ineq{E_{spk}} for a simple SNN.}
		\label{fig:energy_calculation_example}
		\vspace{-10pt}
	\end{center}
\end{figure}

Figure~\ref{fig:energy_calculation_example} shows a simple 2-input and 1-output SNN. The neurons are shown as grayed circles, while the synaptic weights are shown on each connection. The number on a link represents the number of spikes for a given input.

The energy for the 5 spikes generated from the top pre-synaptic neuron = \ineq{5\cdot \left[e_{neuron} + I_{prog}^2\cdot t_{spk}\cdot\left(R_{ON} + \frac{1}{w_1} \right)\right]}. The energy for the 3 spikes from the bottom pre-synaptic neuron = \ineq{3\cdot \left[e_{neuron} + I_{prog}^2\cdot t_{spk}\cdot\left(R_{ON} + \frac{1}{w_2} \right)\right]}. Finally, the energy for the 2 spikes generated from the post-synaptic neuron is \ineq{2\cdot e_{neuron}}. Therefore, the total spike energy is

\begin{footnotesize}
\begin{eqnarray*}
    \label{eq:snn_example_energy}
    \footnotesize E_{spk} &=& 5\cdot \left[e_{neuron} + I_{prog}^2\cdot t_{spk}\cdot\left(R_{ON} + \frac{1}{w_1} \right)\right] +\\
    &&3\cdot \left[e_{neuron} + I_{prog}^2\cdot t_{spk}\cdot\left(R_{ON} + \frac{1}{w_2} \right)\right] + \\
    && 2\cdot e_{neuron}
\end{eqnarray*}
\end{footnotesize}

\begin{figure}[h!]
	\begin{center}
		\includegraphics[width=0.89\columnwidth]{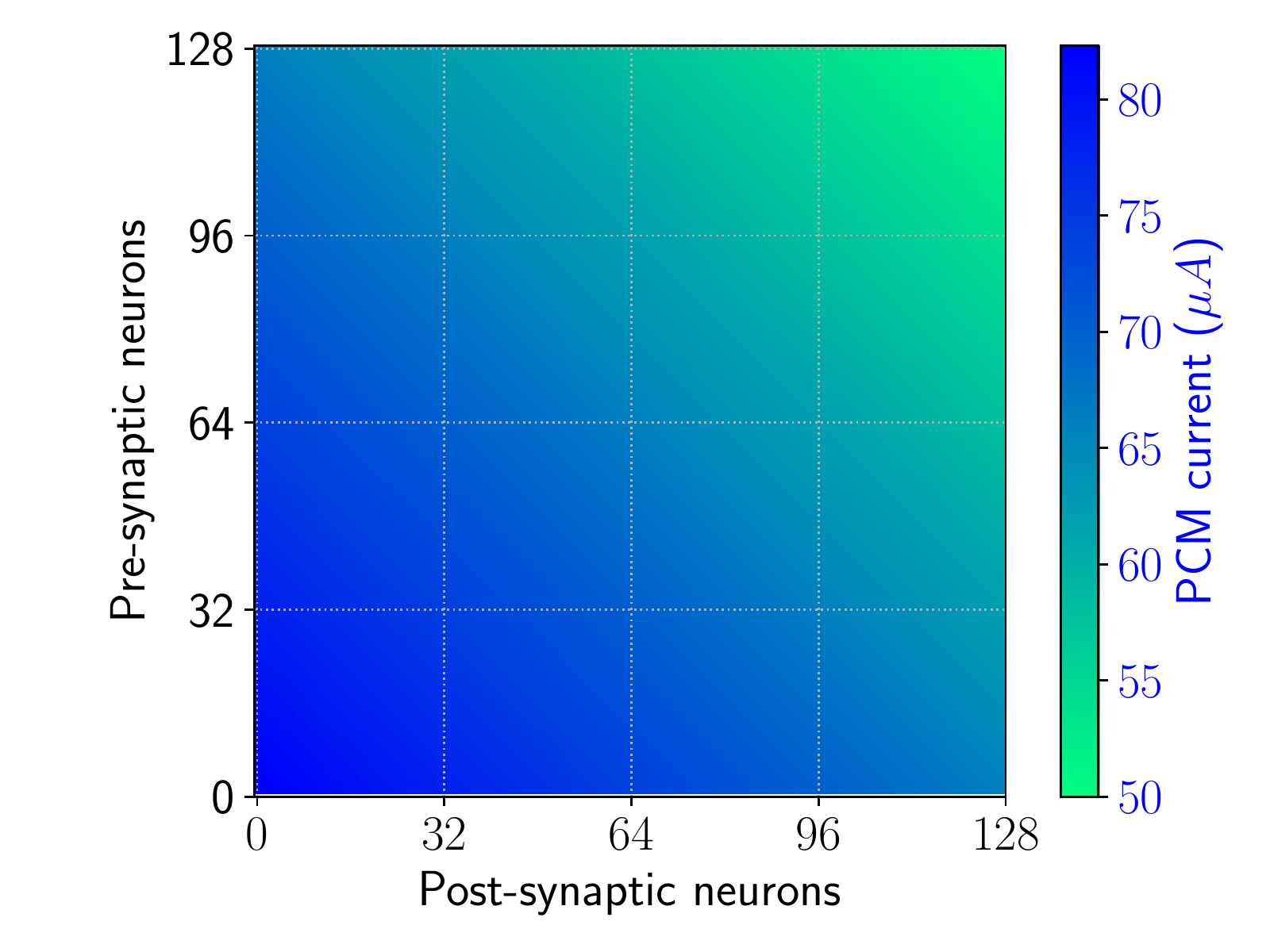}
		\vspace{-10pt}
		\caption{Current map in a 128x128 crossbar.}
		\label{fig:current_map}
	\end{center}
\end{figure}

From a crossbar perspective, the parasitic components on the rows and columns create asymmetry in current propagating through different NVM cells in the crossbar. Figure~\ref{fig:current_map} shows the current variation in a 128x128 PCM crossbar.  

\begin{figure}[h!]
	\begin{center}
		\includegraphics[width=0.99\columnwidth]{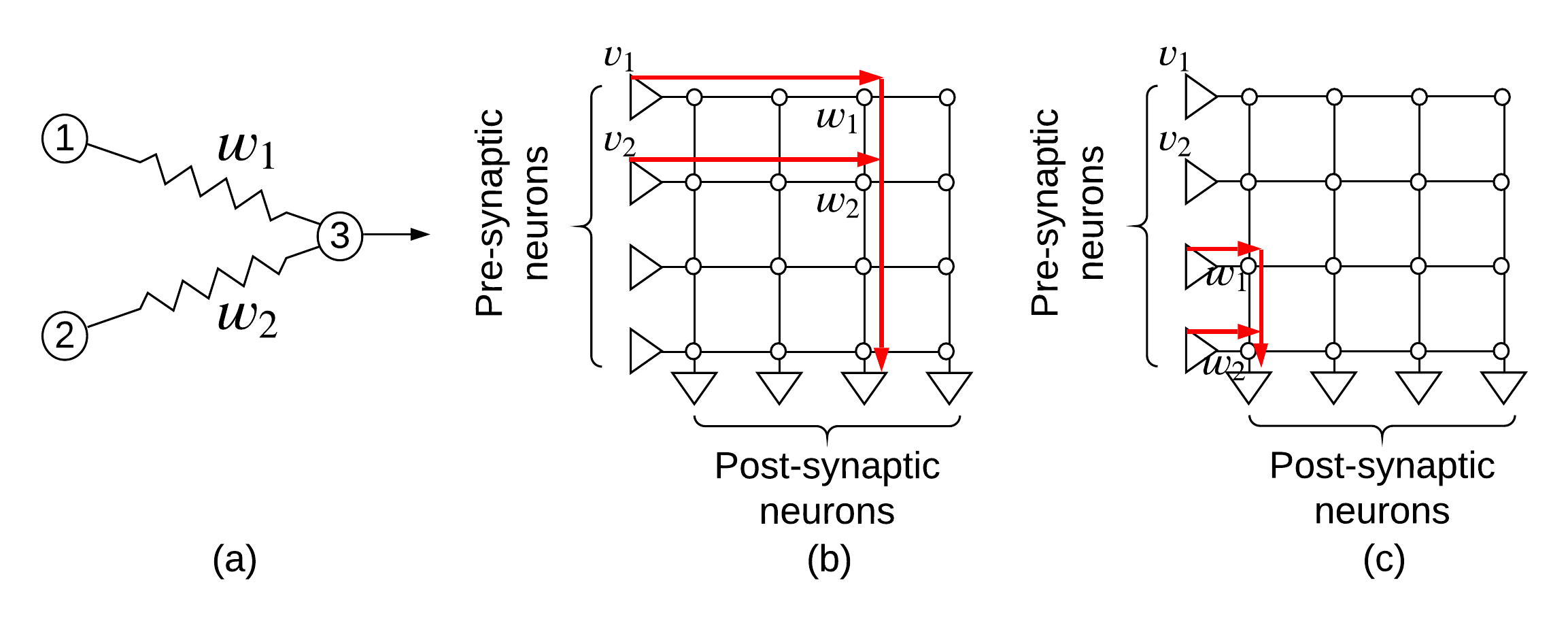}
		\vspace{-10pt}
		\caption{(a) A simple spiking neural network, (b) Mapping of the network to a crossbar, (c) A different mapping of the network to the hardware.}
		\label{fig:mapping_example}
	\end{center}
\end{figure}

Considering these current variations in a crossbar, the programming current \ineq{I_{prog}} is not a constant value for every spike generated in a crossbar. In fact, the programming current is higher for spikes propagating through a synaptic cell located at the bottom left corner than through a synaptic cell located at the top right corner (see Fig.~\ref{fig:current_map}). This is illustrated in Figure~\ref{fig:mapping_example}, which shows two different ways of mapping the SNN of Figure~\ref{fig:mapping_example}a to the crossbar. For Figure~\ref{fig:mapping_example}c, the programming current is higher than the mapping of Figure~\ref{fig:mapping_example}b.

Therefore, the spike energy for an SNN application on a neuromorphic hardware is
\begin{equation}
    \label{eq:spike_energy}
    \footnotesize\boxed{E_{spk} = \sum_{i=0}^{|\textbf{C}|-1} \cdot \sum_{j=0}^{S(C_i)-1}
    \left[e_{neuron} + I^2_{prog}(i,j)\cdot t_{spk}\cdot\left(R_{ON} + \frac{1}{w(i,j)}\right)\right]}
\end{equation}
where \ineq{I_{prog}(i,j)} is the current of the \ineq{j^\text{th}} spike in crossbar \ineq{C_i} and it depends on where the corresponding synaptic connection is mapped within a crossbar.  
\subsection{Communication Energy}
This is the total energy consumed by all spikes on the interconnect of a neuromorphic hardware for a given SNN application working on a representative data.

In mapping clusters to tiles, the inter-cluster spikes are the ones that are communicated over the interconnect. 
Using the formalism of Section~\ref{sec:operational_semantics}, the communication energy is
\begin{equation}
    \footnotesize E_{comm} = \sum_{k=0}^{|\textbf{L}|-1} Spk(L_k)\cdot e_{comm}(L_k)
\end{equation}
where \ineq{e_{comm}(L_k)} is the energy to communicate a spike on the link between the source and destination clusters of the link \ineq{L_k\in \textbf{L}}. \ineq{e_{comm}(L_k)} depends on the hardware architecture and how tiles are interconnected.
In general,
\begin{equation}
    \footnotesize e_{comm}(L_k) = e_{switch}\cdot\left(h_k - 1\right) + e_{wire}\cdot h_k,
\end{equation}
where \ineq{h_k} is the hop distance between the source and destination tiles of the connection \ineq{L_k}, \ineq{e_{wire}} is the energy consumed on the interconnect wires, and \ineq{e_{switch}} is the energy consumed on the switch~\cite{das2012fault,das2013energy,das2013communication,shafik2015adaptive,das2014communication,das2012faultRSP}.
In the following, we consider a mesh-based organization of tiles with X-Y routing of spikes. For this interconnect architecture, the hop count between two tiles is their Manhattan Distance. Specifically, if the source cluster of the link \ineq{L_k}, represented as \ineq{Src(L_k)}, is placed on a tile located at coordinate \ineq{\Big(x_{src}(L_k),y_{src}(L_k)\Big)}, while the destination cluster, represented as \ineq{Dst(L_k)}, is placed on a tile located at coordinate \ineq{\Big(x_{dst}(L_k),y_{dst}(L_k)\Big)}, then

\begin{footnotesize}
\begin{eqnarray}
    \footnotesize h_k &=& ManhattanDistance\Big(Src(L_k),Dst(L_k)\Big) \\
    &=& \Big|x_{src}(L_k)-x_{dst}(L_k)\Big| + \Big|y_{src}(L_k)-y_{dst}(L_k)\Big|\nonumber
\end{eqnarray}
\end{footnotesize}

\begin{figure}[h!]%
    \centering
    \vspace{-10pt}
    \subfloat[Example clustered SNN\label{fig:snn_ex}.]{{\includegraphics[width=3cm]{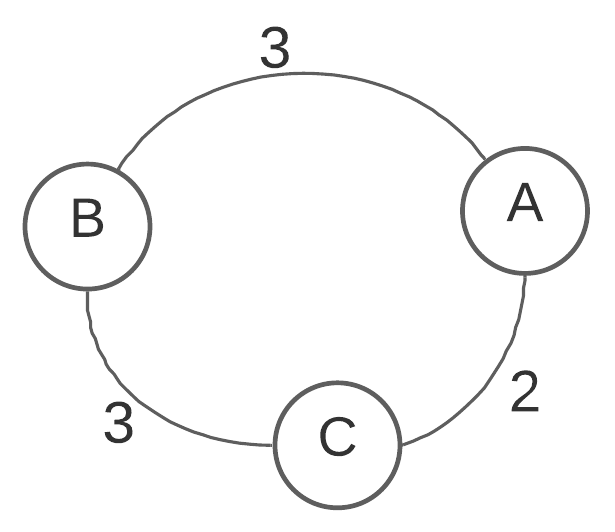} }}%
    \qquad
    \subfloat[Placement of the SNN to a mesh architecture.\label{fig:placement_ex}.]{{\includegraphics[width=4cm]{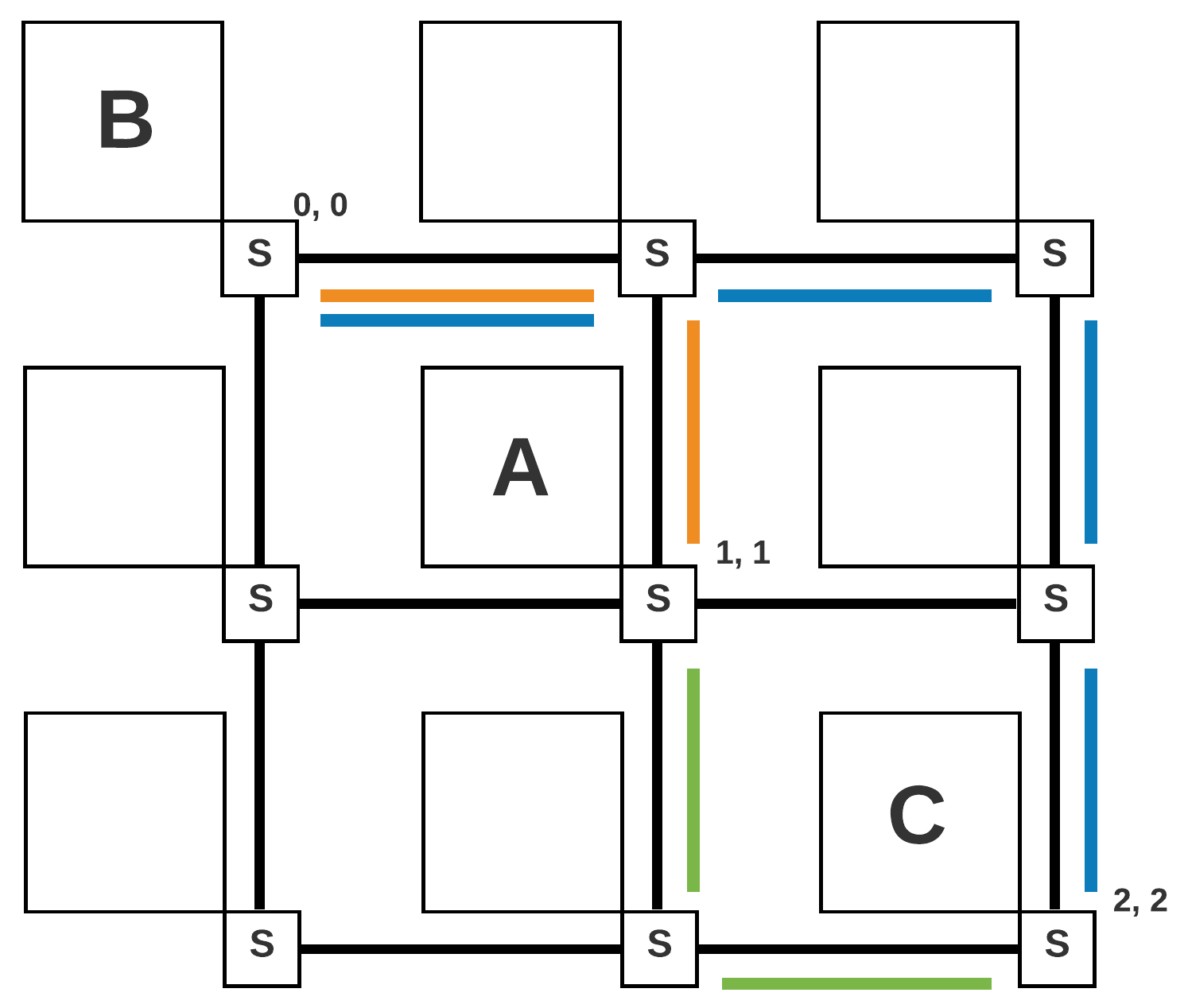} }}%
    \caption{Example of calculation \ineq{E_{comm}} for a clustered SNN placed on a mesh architecture.}%
    \label{fig:comm_energy_compute_ex}%
\end{figure}

Figure~\ref{fig:comm_energy_compute_ex} illustrates the placement of an example clustered SNN to a mesh architecture. Cluster A in this example is placed at coordinate (1,1), cluster B at (0,0), and cluster C at (2,2). As can be seen from this figure, the hop distance between A and B is 2, between B and C is 4, and between C and A is 2. Therefore, the communication energy for spikes communicated between A and B = \ineq{3\cdot \Big[e_{switch} + 2*e_{wire}\Big]}, that between B and C =  \ineq{3\cdot \Big[3\cdot e_{switch} + 4*e_{wire}\Big]}, and that between A and C = \ineq{2\cdot \Big[e_{switch} + 2*e_{wire}\Big]}.

Therefore, the total communication energy is

\begin{footnotesize}
\begin{eqnarray*}
     E_{comm} &=& 3\cdot \Big[e_{switch} + 2*e_{wire}\Big] + \\
     && 3\cdot \Big[3\cdot e_{switch} + 4*e_{wire}\Big] + \\
     && 2\cdot \Big[e_{switch} + 2*e_{wire}\Big]
\end{eqnarray*}
\end{footnotesize}

Overall, the communication energy for an SNN application mapped to a neuromorphic hardware is
\begin{equation}
    \label{eq:communication_energy}
    \footnotesize\boxed{E_{comm} = \sum_{k=0}^{|\textbf{L}|-1} Spk(L_k)\cdot \Big[e_{switch}\cdot\left(h_k - 1\right) + e_{wire}\cdot h_k\Big]}
\end{equation}


\subsection{Energy Dependencies and the Role of System Software in Energy Consumption}\label{sec:motivation}
From Equations \ref{eq:spike_energy} and \ref{eq:communication_energy}, we can conclude that energy consumption depends on 1) how an SNN model is partitioned into clusters (determines the number of neurons and synapses in each cluster), 2) how the clusters are mapped to the neurosynaptic cores (determines the hop distances), and 3) how the neurons and synapses of a cluster are placed inside each core (determines the spikes propagation). All these factors can be controlled via a system software such as \sm{}~\cite{spinemap} and \esl{}~\cite{esl20}. 

\begin{figure}[h!]%
    \centering
    \vspace{-10pt}
    \subfloat[Spike Energy ($E_{spk}$) of \sm{} and \esl{}.\label{fig:motivation_spike_energy}]{{\includegraphics[width=9cm]{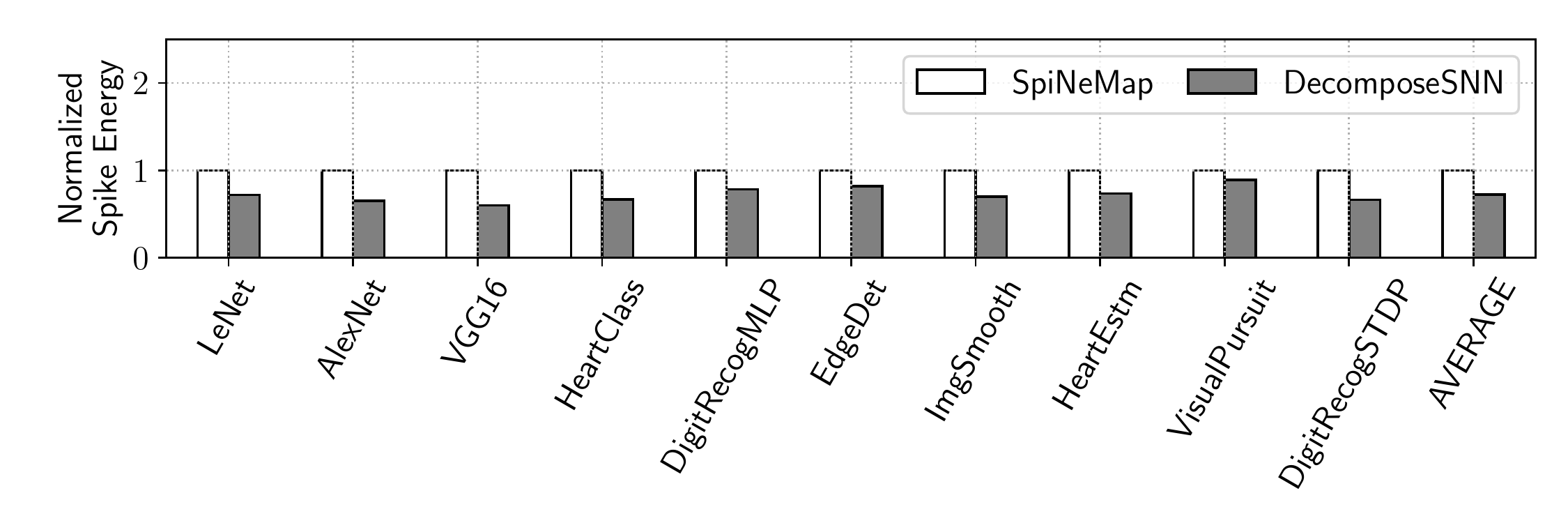} }}%
    \qquad
    \subfloat[Communication Energy ($E_{comm}$) of \sm{} and \esl{}.\label{fig:motivation_comm_energy}]{{\includegraphics[width=9cm]{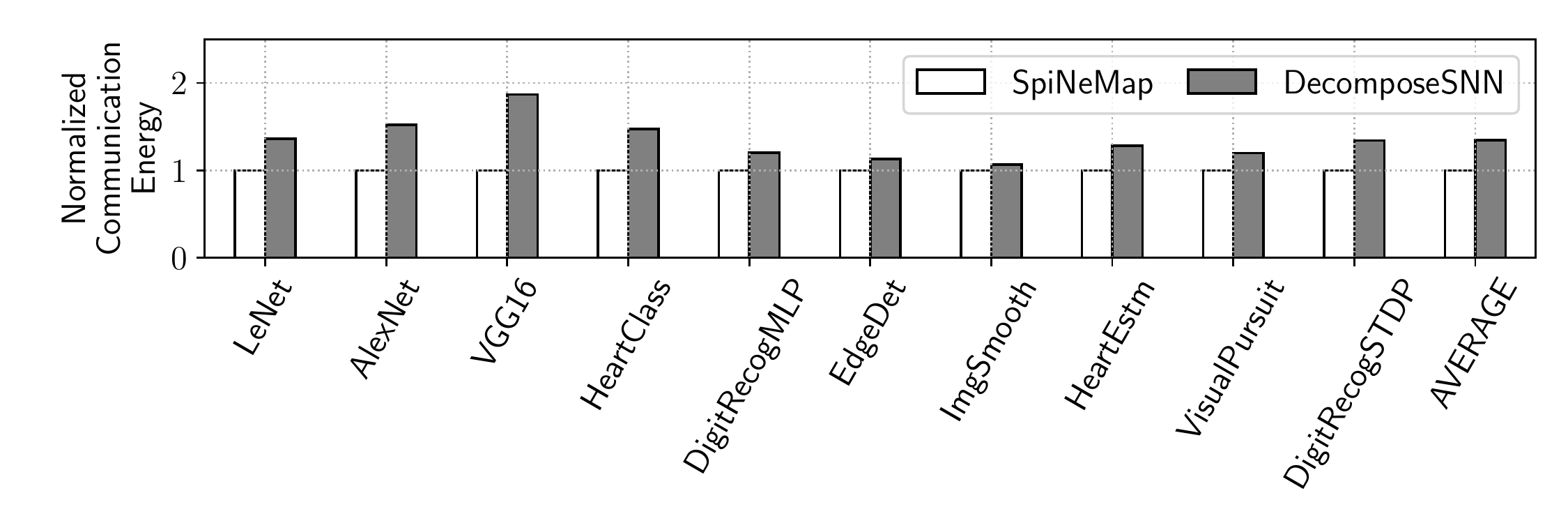} }}%
    \caption{Role of system software in energy management of neuromorphic computing.}%
    \label{fig:system_software_motivation}%
\end{figure}

Figure~\ref{fig:system_software_motivation} compares \sm{}, which minimizes communication on the interconnect and \esl{}, which maximizes crossbar utilization for 10 machine learning applications (see our evaluation methodology in Section~\ref{sec:evaluation}).

We observe that \sm{} has 24\% lower communication energy and 40\% higher spike energy than \esl{}. This is because \sm{} explicitly minimizes spike communication on the interconnect and therefore, has lower communication energy. On the other hand, \esl{} maximizes crossbar utilization and therefore, generates fewer clusters than \sm{}, resulting in lower spike energy. These results are consistent with the reported results in the two corresponding publications.

\begin{figure}[h!]
	\begin{center}
		\includegraphics[width=0.99\columnwidth]{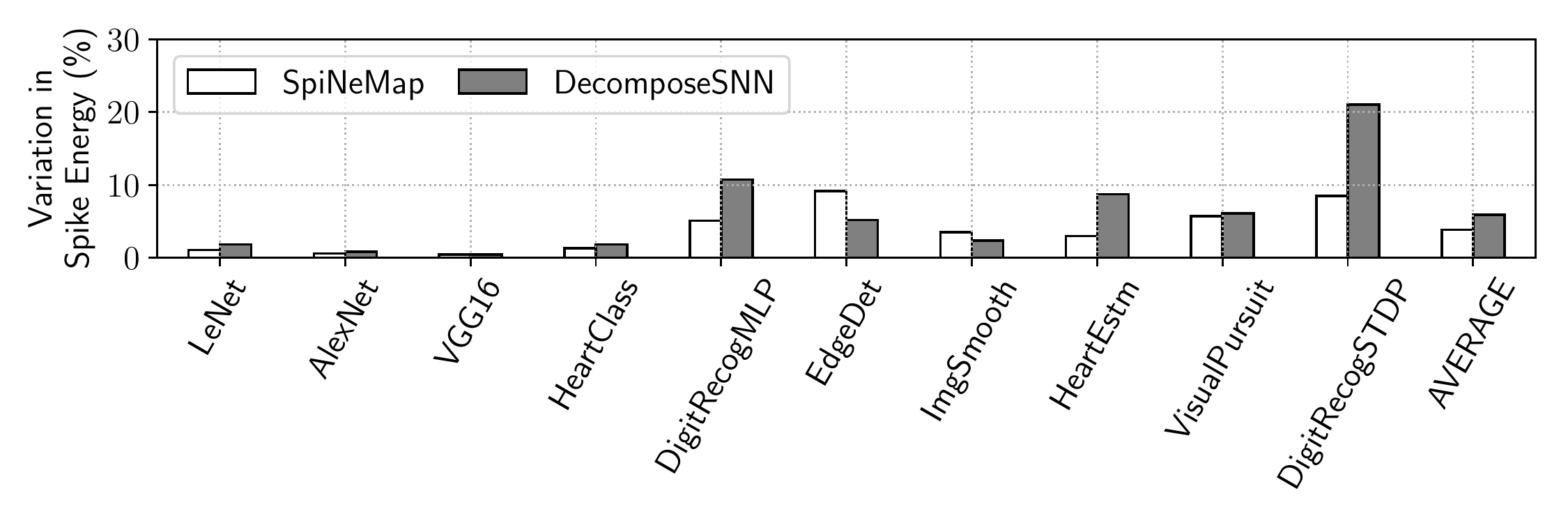}
		\vspace{-10pt}
		\caption{Variation in spike energy due to different synapse placement strategies on crossbars.}
		\label{fig:motivation_variation}
	\end{center}
\end{figure}

To highlight the importance of neuron and synapse placement within each crossbar (see our motivation example in Figure~\ref{fig:mapping_example}), Figure~\ref{fig:motivation_variation} shows the variation between minimum and maximum spike energy for \sm{} and Decompose\-SNN considering 100 random placements of synapses of clusters to the crossbars of a neuromorphic hardware.

We observe that the spike energy of \sm{} varies by 3.8\% and that of \esl{} varies by 5.9\% depending on how synapses are placed on crossbars.

We conclude that the system software of a neuromorphic hardware can play a key role in managing the energy consumption of neuromorphic computing.

\section{Energy-Aware System Software}\label{sec:system_software_energy}
Using the motivation presented in Section~\ref{sec:motivation}, we now present our energy-aware system software to map machine learning applications to neuromorphic hardware.

\begin{figure}[h!]
	\begin{center}
		\vspace{-10pt}
		\includegraphics[width=0.99\columnwidth]{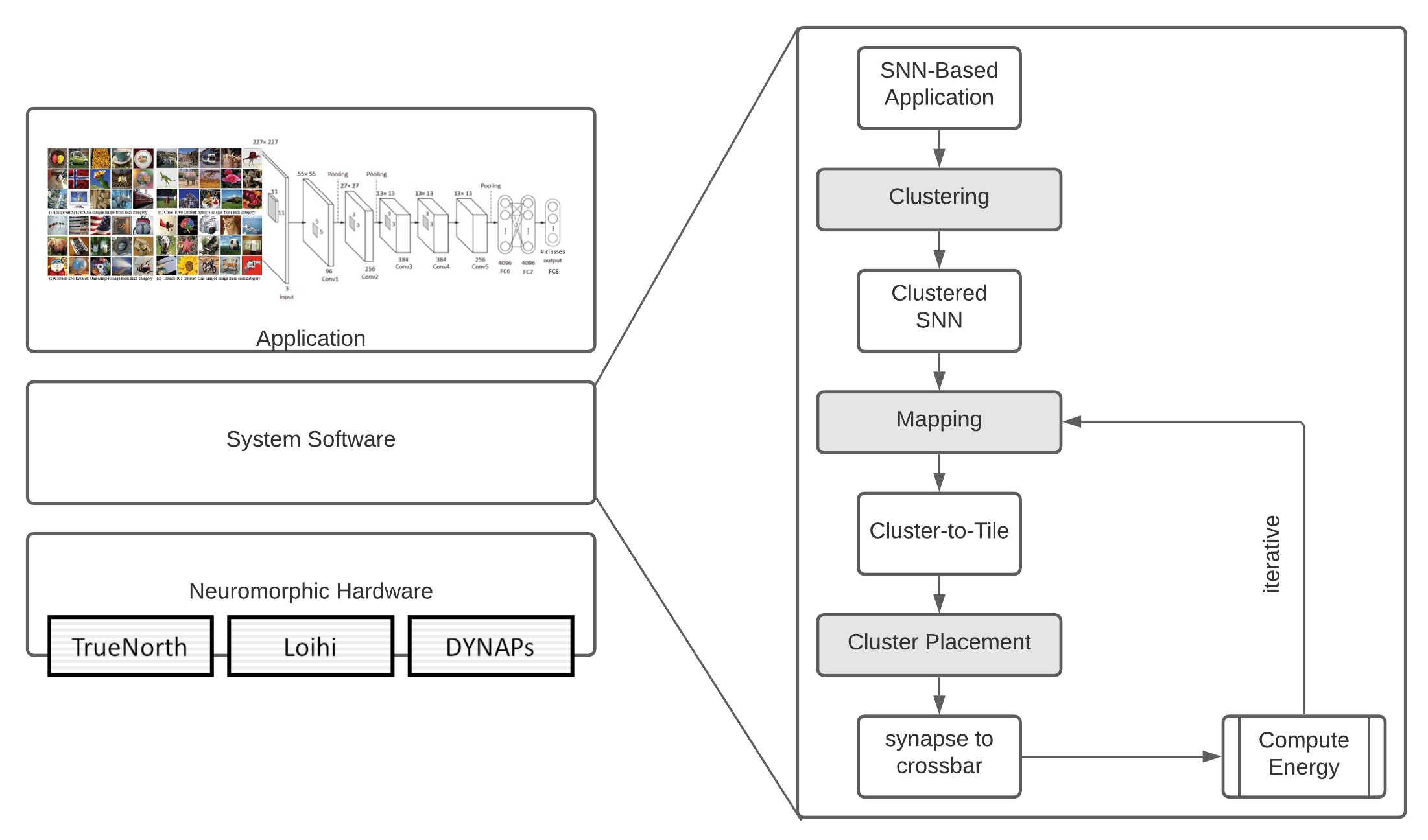}
		\vspace{-10pt}
		\caption{Our energy-aware system software.}
		\label{fig:system_software}
		\vspace{-10pt}
	\end{center}
\end{figure}

Figure~\ref{fig:system_software} shows a neuromorphic system comprising of the application layer, the system software layer, and the hardware layer. The application layer at the top consists of the user space to run machine learning applications. In this illustration we show the execution of AlexNet for ImageNet classification. The hardware layer at the bottom consists of the neuromorphic hardware such as TrueNorth~\cite{truenorth}, Loihi~\cite{loihi}, and DYNAPs~\cite{dynapse}. At the middle is the system software layer, which interacts with both the application and hardware layers. The system software performs energy optimization using the iterative approach shown to the right. 

The workflow of the system software involves clustering a machine learning application to generate clustered SNN graph. Next, the clusters are mapped to the tiles of the hardware using a mapping approach. Finally, the clusters are placed to crossbars using the placement step. Although the clustering step could potentially be incorporated inside the iterative loop, we placed it outside to limit the complexity of the design space exploration. In fact, clustering of applications is an NP-hard problem as shown in \sm{}~\cite{spinemap}. Our clustering approach uses the graph partitioning algorithm of \sm{}, minimizing 1) inter-cluster communication (similar to \sm{}, and 2) maximizing cluster utilization (similar to \esl{}).

We now describe the iterative approach of energy minimization starting from a clustered SNN using Algorithm~\ref{alg:min_temp}. At the heart of this algorithm is the \texttt{CalculateEnergy} function, which calculates the total energy consumption using Equations~\ref{eq:spike_energy} and \ref{eq:communication_energy}. The \texttt{AssignCluster} function is a greedy heuristic to place a cluster to a crossbar. For this purpose, we first sort (in descending order) the synapses of a cluster in terms of their number of spikes. Next, the synapses are allocated to the crossbars, ensuring that the most activated one is placed towards the top right corner, where the spike current is lower.

\begin{algorithm}[h]
	\scriptsize{
	    \KwIn{$G_\text{CSNS},G_\text{NH}$}
	    \KwOut{${M}$}
	    \For{$i$ in $MaxIter$}{
	        ${M}_\text{init}$ = allocate clusters to tiles randomly\;
	        AssignCluster()\;
	        $E_\text{init}$ = \texttt{CalculateEnergy}(${M}_\text{init}$)\;
	        \For{$C_x,C_y\in \textbf{C}$}{
	            $M = M_\text{init}$\;
	            Find $T_i,T_j$ such that $m_{x,i} = m_{y,j} = 1$ \tcc{Find the tiles to which $C_x$ and $C_y$ are mapped.}
	            Change $M$ to set $m_{x,j} = m_{y,i} = 1$ and $m_{x,i} = m_{y,j} = 0$\tcc{Swap the tiles of $C_x$ and $C_y$.}
	            AssignCluster()\;
	            $E = \texttt{CalculateEnergy}(M)$\tcc{Calculate energy of the new mapping.}
	            \If{$E < E_\text{init}$}{
	                $M_\text{init} = M$ and $M_\text{min} = M$\tcc{If energy reduces then retain the new mapping.}
	            }
	        }
	    }
	    Return ${M}_\text{min}$
	}
	\caption{\small Place neuron and synapse to neuromorphic hardware, minimizing the energy consumption.}
	\label{alg:min_temp}
\end{algorithm}

Algorithm~\ref{alg:min_temp} proceeds as follows. First, the clusters are randomly allocated to tiles (line 2). Next, the energy consumption is computed after assigning the synapses to the crossbars (lines 2-3). Then, for every cluster pair, the algorithm swaps their tile and compute the new energy (lines 6-10). If the energy improves, then the swapping is retained and the algorithm proceeds to the next cluster pair (lines 11-13). The algorithm is iterated for \ineq{MaxIter} iterations, each time starting from a new random allocation of clusters (lines 1-15). The minimum energy mapping is returned.

In this algorithm, \ineq{MaxIter} is a user-defined parameter and is used to explore the trade-offs between mapping time and the solution quality (see Section~\ref{sec:evaluation}).

\section{Evaluation}\label{sec:evaluation}
We evaluated 10 machine learning applications that are representative of three most commonly used neural network classes --- convolutional neural network (CNN), multi-layer perceptron (MLP), and recurrent neural network (RNN).
These applications are summarized in Table~\ref{tab:apps}. We simulate these applications on an in-house cycle-accurate neuromorphic hardware simulator. We model the DYNAPs neuromorphic hardware~\cite{dynapse} with the following configurations.

\begin{table}[h!]
	\renewcommand{\arraystretch}{0.8}
	\setlength{\tabcolsep}{2pt}
	\caption{Evaluated applications.}
	\label{tab:apps}
	\centering
	\begin{threeparttable}
	{\fontsize{6}{10}\selectfont
		\begin{tabular}{cc|ccl|c}
			\hline
			\textbf{Class} & \textbf{Applications} & \textbf{Synapses} & \textbf{Neurons} & \textbf{Topology} & \textbf{Accuracy}\\
			\hline
			\multirow{4}{*}{CNN} & LeNet~\cite{lenet} & 282,936 & 20,602 & CNN & 85.1\%\\
			& AlexNet~\cite{alexnet} & 38,730,222 & 230,443 & CNN & 90.7\%\\
			& VGG16~\cite{vgg16} & 99,080,704 & 554,059 & CNN & 69.8 \%\\
			& HeartClass~\cite{jolpe18} & 1,049,249 & 153,730 & CNN & 63.7\%\\
			\hline
			\multirow{3}{*}{MLP} & DigitRecogMLP & 79,400 & 884 & FeedForward (784, 100, 10) & 91.6\%\\
			& EdgeDet \cite{carlsim} & 114,057 &  6,120 & FeedForward (4096, 1024, 1024, 1024) & 100\%\\
			& ImgSmooth \cite{carlsim} & 9,025 & 4,096 & FeedForward (4096, 1024) & 100\%\\
			\hline
 			\multirow{3}{*}{RNN} & HeartEstm \cite{HeartEstmNN} & 66,406 & 166 & Recurrent Reservoir & 100\%\\
 			& VisualPursuit \cite{Kashyap2018} & 163,880 & 205 & Recurrent Reservoir & 47.3\%\\
 			& DigitRecogSTDP \cite{Diehl2015} & 11,442 & 567 & Recurrent Reservoir & 83.6\%\\
			\hline
	\end{tabular}}
	\end{threeparttable}
\end{table}

\begin{itemize}
    \item A tiled array of 4 tiles, each with a 128x128 crossbar. There are 65,536 synapses per crossbar.
    \item Spikes are digitized and communicated between cores through a mesh routing network using the Address Event Representation (AER) protocol.
    \item Each synaptic element is a PCM cell implementing multi-bit synapse. 
\end{itemize}

Table \ref{tab:hw_parameters} reports the hardware parameters of DYNAP-SE.

\begin{table}[h!]
    \caption{Major simulation parameters extracted from \cite{dynapse} and extrapolated for PCM technology.}
	\label{tab:hw_parameters}
	\centering
	{\fontsize{6}{10}\selectfont
		\begin{tabular}{lp{5cm}}
			\hline
			Neuron technology & 65nm CMOS\\
			\hline
			Synapse technology & PCM\\
			\hline
			Supply voltage & 1.0V\\
			\hline
			$E_{neuron}$ & 50pJ at 30Hz spike frequency\\
			\hline
			$e_{switch} + 2*e_{wire}$ & 147pJ\\
			\hline
			Switch bandwidth & 1.8G. Events/s\\
			\hline
	\end{tabular}}
\end{table}

\subsection{Energy Consumption}
\begin{figure}[h!]
	\begin{center}
		\vspace{-10pt}
		\includegraphics[width=0.99\columnwidth]{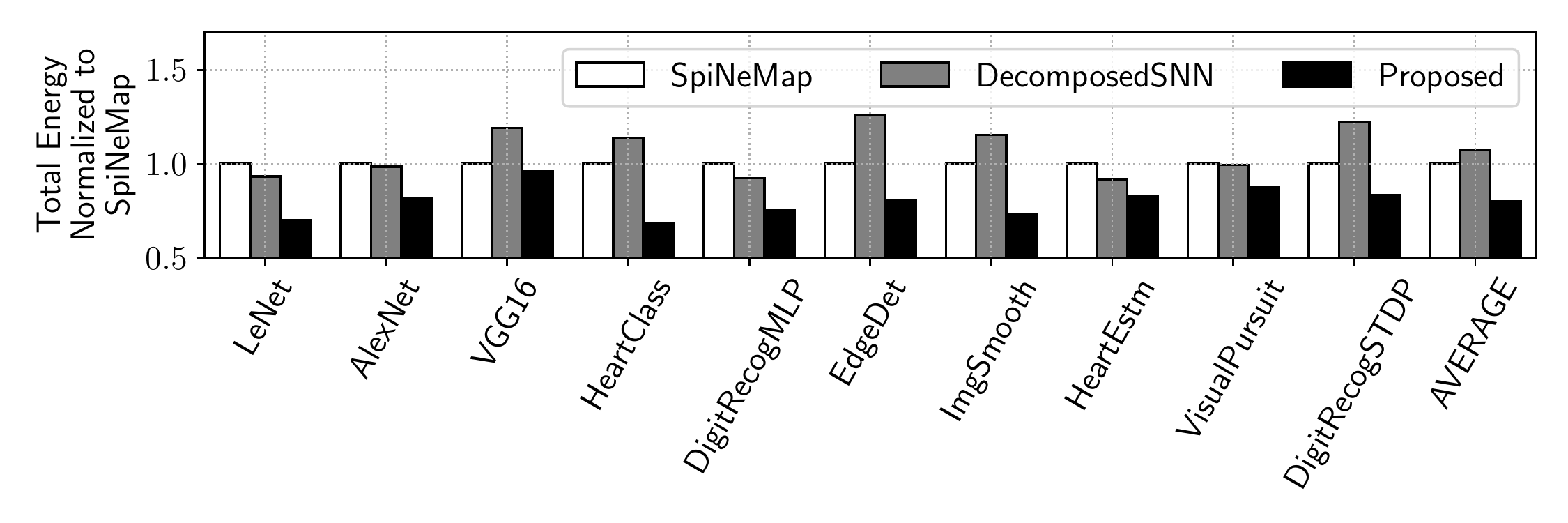}
		\vspace{-10pt}
		\caption{Total energy normalized to \sm{}.}
		\label{fig:energy}
		\vspace{-10pt}
	\end{center}
\end{figure}

Figure~\ref{fig:energy} reports the total energy consumption for each application for the evaluated approaches normalized to \sm{}. We make the following two observations. 

First, between \sm{} and \esl{}, the energy consumption of \sm{} is lower than \esl{} for applications such as VGG16, HeartClass, and EdgeDet. For these applications, there is a significant number of spikes communicated on the interconnect. Therefore, by reducing inter-cluster communication, \sm{} also reduces energy consumption. For other applications such as LeNet and HeartEstm, the number of inter-cluster spikes is less, so \esl{}, which maximizes cluster utilization improves on the total energy consumption. Second, compared to both these approaches, the proposed approach results in 20\% lower energy than \sm{} and 24\% lower energy than \esl{}. The improvement over both these approaches is because the proposed approach explicitly incorporates both the spike and communication energy in finding a suitable mapping of clusters to the hardware.

\begin{figure}[h!]
	\begin{center}
		\vspace{-10pt}
		\includegraphics[width=0.99\columnwidth]{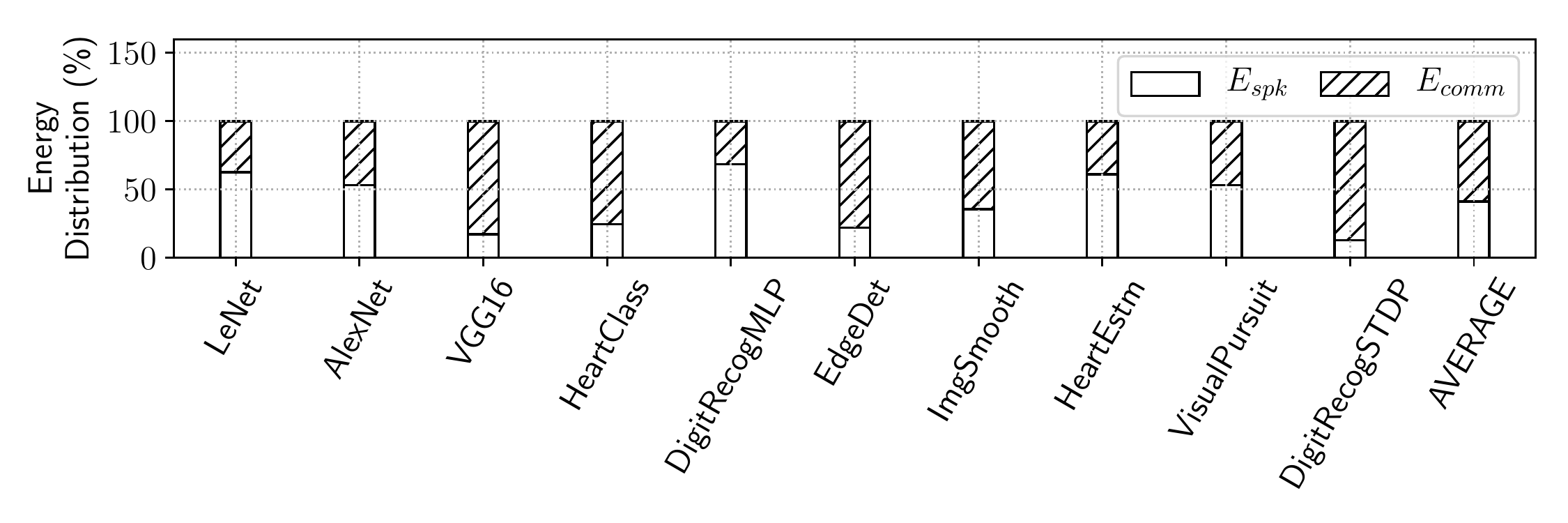}
		\vspace{-10pt}
		\caption{Total energy distributed into $E_{spk}$ and $E_{comm}$.}
		\label{fig:energy_split}
		\vspace{-10pt}
	\end{center}
\end{figure}

To give further insight, Figure~\ref{fig:energy_split} reports the total energy, distributed into spike energy (\ineq{E_{spk}}) and communication energy (\ineq{E_{comm}}). We observe that communication energy constitute an average 58.8\% of the total energy consumption and it depends on the total spikes generated in a workload.

\subsection{Spike Latency and Model Performance}
\begin{figure}[h!]
	\begin{center}
		\vspace{-10pt}
		\includegraphics[width=0.99\columnwidth]{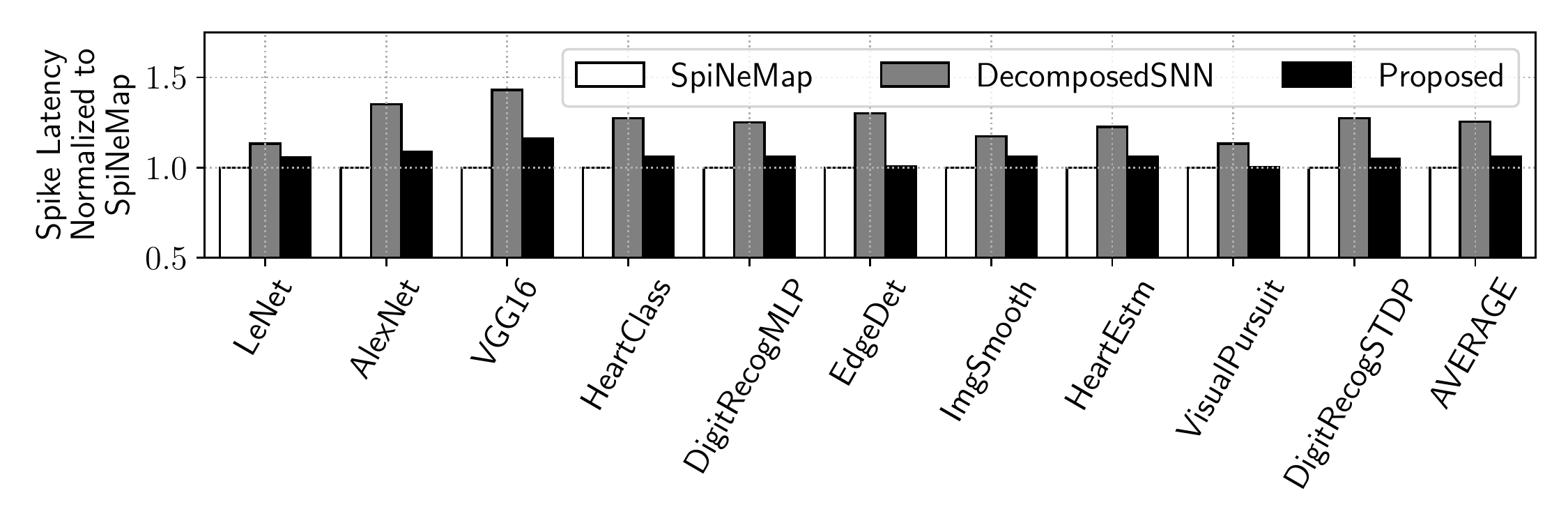}
		\vspace{-10pt}
		\caption{Spike latency normalized to \sm{}.}
		\label{fig:latency}
		\vspace{-10pt}
	\end{center}
\end{figure}

Figure~\ref{fig:latency} plots the spike latency for each evaluated applications for the evaluated approaches normalized to \sm{}. We make the following two observations.

First, \sm{} has the lowest latency. This is because \sm{} minimizes spike congestion on the interconnect, which reduces spike delay. \esl{} has the highest latency because of its optimization objective, which is to maximize utilization. Second, the proposed approach minimizes spike communication to reduce the communication energy. This also reduces the spike latency. Overall, the spike latency of the proposed approach is only 6\% higher than \sm{} but 15\% lower than \esl{}.
As described in \cite{spinemap}, spike latency can lead to a loss in model performance. Therefore, by keeping the spike latency reasonably close to \sm{}, the performance of the proposed approach is similar to that reported in column 6 of Table~\ref{tab:apps}.


\subsection{Architecture Evaluation}
\begin{figure}[h!]
	\begin{center}
		\vspace{-10pt}
		\includegraphics[width=0.99\columnwidth]{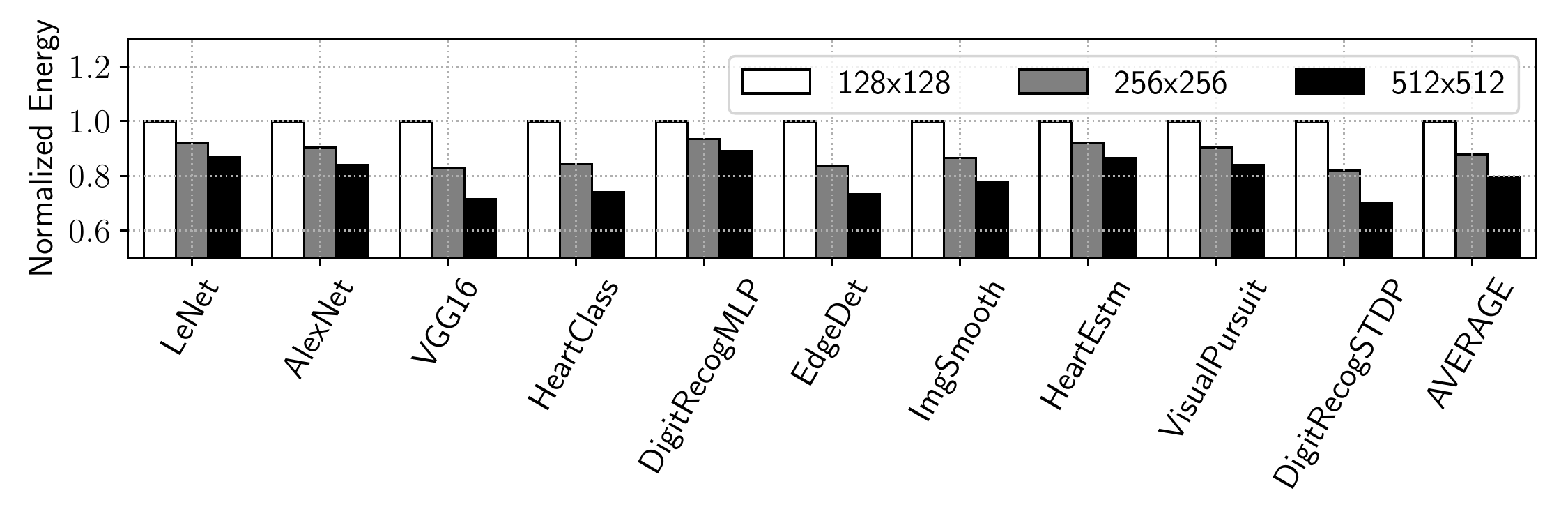}
		\vspace{-10pt}
		\caption{Energy consumption for different hardware configurations.}
		\label{fig:tiles}
		\vspace{-10pt}
	\end{center}
\end{figure}

Figure~\ref{fig:tiles} report the energy consumption for three different hardware configurations -- with 128x128, 256x256, and 512x512 crossbars, for the evaluated applications. Results are normalized to the energy consumption with 128x128 crossbars. We observe that the energy consumption is 13\% and 28\% lower when using 256x256 and 512x512 crossbars compared to using 128x128 crossbars. Energy consumption reduces when using larger crossbars because of the reduction in the total number of spikes on the interconnect.

\subsection{Solution Quality Evaluation}
Table~\ref{tab:compile_time} reports the mapping time and the normalized energy obtained for three different settings of the parameter $MaxIter$. We observe that as $MaxIter$ is increased, the energy consumption reduces for all applications. This is because with the increase in the number of iterations, Algorithm~\ref{alg:min_temp} is able to find a better solution. However, the mapping time also increases. Finally, we observe that increasing $MaxIter$ from 100 to 1000 results in a significant increase in mapping time with a minimal improvement of the total energy. We conclude that setting $MaxIter = 100$ gives the best trade-off in terms of mapping time and the solution quality.
Users can use this $MaxIter$ parameter to set a limit on the time of their algorithm by analyzing the complexity of their model against the ones we evaluate (see Table~\ref{tab:apps}).

\begin{table}[h!]
	\renewcommand{\arraystretch}{1.2}
	\setlength{\tabcolsep}{1.5pt}
	\caption{Mapping time and solution trade-off.}
	\label{tab:compile_time}
	\centering
	{\fontsize{6}{9}\selectfont
		\begin{tabular}{r|c|c|c|c|c|c}
			\hline
			\multirow{3}{*}{\textbf{Application}} & \multicolumn{2}{|c|}{$\mathbf{MaxIter = 10}$} & \multicolumn{2}{|c|}{$\mathbf{MaxIter = 100}$} & \multicolumn{2}{|c}{$\mathbf{MaxIter = 1000}$}\\
			\cline{2-7}
			& \textbf{Mapping} & \textbf{Total} & \textbf{Mapping} & \textbf{Total} & \textbf{Mapping} & \textbf{Total}\\
			& \textbf{Time (sec)} & \textbf{Energy} & \textbf{Time (sec)} & \textbf{Energy} & \textbf{Time (sec)} & \textbf{Energy}\\
			\hline
			LeNet & 75 & 1.13 & 321 & 1.0 & 2700	& 0.99\\
            AlexNet & 200 & 1.1 & 2189 & 1.0 & 12008 & 1.0\\
            VGG16 & 241 & 1.06 & 2989 & 1.0 & 34300 & 1.0\\
            HeartClass & 116 & 1.16 & 1008 & 1.0 & 10178 & 1.03\\
            DigitRecogMLP & 10 & 1.17 & 160 & 1.0 & 1600 & 0.97\\
            EdgeDet & 25 & 1.15 & 200 & 1.0 & 1898 & 0.98\\
            ImgSmooth & 50 & 1.11 & 400 & 1.0 & 1940 & 0.99\\
            HeartEstm & 15 & 1.08 & 156 & 1.0 & 1344 & 0.97\\
            VisualPursuit & 30 & 1.0 & 324	 & 1.0 & 3003 & 1.0\\
            DigitRecogSTDP & 15 & 1.07 & 164 & 1.0 & 1615 & 0.9\\
    \hline
	\end{tabular}}
\end{table}


\section{Conclusion}\label{sec:conclusions}
In this work, we provide a comprehensive energy model for executing machine learning applications on neuromorphic hardware. 
Using this model we show that the system software for neuromorphic hardware plays a critical role in energy management of neuromorphic computing by controlling  1) how an SNN model is partitioned into clusters, 2) how the clusters are mapped to the neurosynaptic cores of the hardware, and 3) how the neurons and synapses of a cluster are placed inside each core.
We then propose a heuristic to perform energy-aware application mapping to neuromorphic hardware, lowering the overall energy consumption. 
Using this heuristic, we show that the energy consumption can be reduced by an average 20\% compared to a state-of-the-art for typical machine learning applications.

\section{Acknowledgments}
This work is supported by 1) the National Science Foundation Award CCF-1937419 (RTML: Small: Design of System Software to Facilitate Real-Time Neuromorphic Computing) and 2) the National Science Foundation Faculty Early Career Development Award CCF-1942697 (CAREER: Facilitating Dependable Neuromorphic Computing: Vision, Architecture, and Impact on Programmability).


\balance
\bibliographystyle{IEEEtranSN}
\bibliography{commands,disco,external}

\end{document}